\documentclass[conference]{IEEEtran}
\IEEEoverridecommandlockouts
\usepackage{cite}
\usepackage{amsmath,amssymb,amsfonts}
\usepackage{algorithmic}
\usepackage{graphicx}
\usepackage{textcomp}
\usepackage{xcolor}
\usepackage{hyperref}
\usepackage{tabularx}
\usepackage{booktabs}
\usepackage{multirow}
\usepackage{placeins}
\usepackage{subfigure}
\usepackage[utf8]{inputenc} 
\usepackage[T1]{fontenc}    
\usepackage{hyperref}       
\usepackage{url}           
\usepackage{booktabs}          
\usepackage{nicefrac}       
\usepackage{microtype}      
\usepackage{lipsum}
\usepackage{graphicx} 
\usepackage{subcaption}
\usepackage{graphicx} 
\usepackage{multirow}
\usepackage{stfloats}
\usepackage{float}     
\usepackage{xcolor} 
\usepackage{attachfile}
\usepackage{pdfpages}

\graphicspath{{img/}}

\def\BibTeX{{\rm B\kern-.05em{\sc i\kern-.025em b}\kern-.08em
    T\kern-.1667em\lower.7ex\hbox{E}\kern-.125emX}}

\begin{document}

\title{EEG-MFTNet: An Enhanced EEGNet Architecture with Multi-Scale Temporal Convolutions and Transformer Fusion for Cross-Session Motor Imagery Decoding\
}

\author{\IEEEauthorblockN{Panagiotis Andrikopoulos, Siamak Mehrkanoon\IEEEauthorrefmark{1}\thanks{*Corresponding author.}\thanks{Code available at: \url{https://github.com/pandrikopoul/EEG-MFTNet}}}
\IEEEauthorblockA{\textit{Department of Information and Computing Sciences, Utrecht University, Utrecht, Netherlands} \\
panosandrikos3@gmail.com, s.mehrkanoon@uu.nl}
}

\maketitle

\begin{abstract}

Brain-computer interfaces (BCIs) enable direct communication between the brain and external devices, providing critical support for individuals with motor impairments. However, accurate motor imagery (MI) decoding from electroencephalography (EEG) remains challenging due to noise and cross-session variability. This study introduces EEG-MFTNet, a novel deep learning model based on the EEGNet architecture, enhanced with multi-scale temporal convolutions and a Transformer encoder stream. These components are designed to capture both short and long-range temporal dependencies in EEG signals. The model is evaluated on the SHU dataset using a subject-dependent cross-session setup, outperforming baseline models, including EEGNet and its recent derivatives. EEG-MFTNet achieves an average classification accuracy of 58.9\% while maintaining low computational complexity and inference latency. The results highlight the model’s potential for real-time BCI applications and underscore the importance of architectural innovations in improving MI decoding. This work contributes to the development of more robust and adaptive BCI systems, with implications for assistive technologies and neurorehabilitation.

\end{abstract}
\section{Introduction}

Brain-computer interfaces (BCIs) enable direct communication between brain activity and external devices, bypassing neuromuscular control to assist individuals with motor impairments \cite{nicolas_alonso2012, altaheri2023deep}. Applications span rehabilitation and assistive robotics, enhancing autonomy through control of prosthetics, wheelchairs, and exoskeletons \cite{meng2025paradigms}. A typical BCI pipeline consists of signal acquisition, feature extraction and classification, followed by command execution \cite{altaheri2023deep}. Brain signals can be recorded using either invasive or non-invasive methods. Non-invasive techniques, such as functional magnetic resonance imaging (fMRI), magnetoencephalography (MEG), and electroencephalography (EEG), are generally safer and more practical for everyday use. While fMRI and MEG offer high spatial resolution, EEG is often preferred for real-time motor imagery (MI) BCIs due to its low cost, portability, and high temporal resolution \cite{ramadan2017brain, sun2025signal}.

MI involves mentally simulating physical movements, which triggers neural activity that can be captured using EEG \cite{altaheri2023deep, padfield2019eeg}. Several EEG rhythms are associated with different cognitive and motor processes, including delta ($\leq$4~Hz), theta (4--8~Hz), alpha and mu (8--13~Hz), beta (13--30~Hz), and gamma ($\geq$31~Hz).  In MI tasks, specific changes in the mu and beta bands, referred to as event-related desynchronization (ERD) and synchronization (ERS), are commonly used as indicators of motor cortex activation \cite{saibene2023eeg, edelman2024non}. However, accurate MI-EEG classification is hindered by signal variability and noise due to emotional, physiological, and environmental factors \cite{padfield2019eeg}. Traditional approaches, such as Common Spatial Patterns (CSP) \cite{ramoser2000optimal} and Filter Bank CSP (FBCSP) \cite{ang2008filter}, extract handcrafted features and typically employ classifiers like linear discriminant analysis (LDA) or support vector machines (SVM). While effective in some settings, these methods require expert-driven parameter tuning and are highly sensitive to both inter-subject and intra-subject variability \cite{li2024mfrc}.

Recent deep learning approaches have improved MI classification by learning features directly from raw EEG, reducing dependence on manual preprocessing and increasing robustness \cite{altaheri2023deep, wang2024depth}. Still, challenges remain in managing cross-session and cross-subject variability without increasing computational complexity. This work addresses these limitations by enhancing an end-to-end model for robust MI-EEG decoding. Specifically, the contributions of this work are as follows:
\begin{itemize}
    \item \textbf{EEG-MFTNet Architecture:} We propose a novel motor imagery EEG decoding model that extends EEGNet by incorporating (i) a multi-scale temporal convolution block to capture patterns across different temporal resolutions, and (ii) a Transformer encoder stream to model long-range temporal dependencies.
    
    \item \textbf{Cross-Session Robustness:} The model is evaluated under a subject-dependent cross-session protocol on the SHU~\cite{ma2022large} dataset, achieving an average accuracy of 58.9\%, which represents a +5\% absolute improvement over the baseline EEGNet, while maintaining low computational cost and real-time inference capability.
    
    \item \textbf{Component Analysis:} An ablation study highlights the complementary benefits of the convolutional and Transformer modules, with each contributing to the overall performance gain.
    
    \item \textbf{Interpretability:} We provide an in-depth interpretability analysis using Gradient $\times$ Input and electrode deletion tests, demonstrating that the model focuses on class-specific electrodes and identifies meaningful neurophysiological patterns, thereby enhancing transparency and trustworthiness.
    
    \item \textbf{Reproducibility:} To support future research, the full implementation of the proposed model and experimental setup is made publicly available, addressing the current lack of reproducible implementations for this dataset.
\end{itemize}


The remainder of this paper is organized as follows. Section~\ref{Related Work} reviews relevant literature on BCI systems and motor imagery classification. Section~\ref{sec:propmodel} describes the proposed model architecture and its principal components. Section~\ref{Data description} details the dataset and preprocessing procedures. Section~\ref{Experiment Results and Discussion} presents the experimental setup, results, performance comparisons, and the ablation study. Section~\ref{Model Interpretation} discusses the interpretability of the proposed model. Finally, Section~\ref{Conclusion and Future work} summarizes the main findings and outlines directions for future research.

\section{Related Work}
\label{Related Work}

MI classification using EEG signals has seen significant progress through deep learning, addressing challenges like inter-subject variability and session-to-session inconsistency. Deep learning models have enabled end-to-end learning directly from raw EEG data. Early models like Deep and Shallow ConvNets~\cite{schirrmeister2017deep} demonstrated the feasibility of CNNs for EEG decoding. EEGNet~\cite{lawhern2018eegnet} further advanced this by introducing a compact architecture using depthwise and separable convolutions. Its success inspired several extensions: EEG-TCNet~\cite{ingolfsson2020eeg} added Temporal Convolutional Networks (TCNs) for long-range modeling, while MI-EEGNet~\cite{riyad2021mi} and EEG-ITNet~\cite{salami2022eeg} introduced multi-scale and inception-style modules. Notably, EEG-ITNet also integrates TCNs with residual connections to enhance temporal feature extraction. To address temporal asymmetry in TCNs, EEG-CDILNet~\cite{liang2023eeg} uses circular dilated convolutions for balanced temporal modeling. EEG-GENet~\cite{wang2022eeg_genet} incorporates graph embeddings to capture spatial relationships between electrodes. Lightweight models like EEG-SimpleConv~\cite{el2024strong} show that simpler architectures can still perform competitively, making them suitable for real-time applications. Other innovations include MFRC-Net~\cite{li2024mfrc}, which balances performance and efficiency via multi-scale residual blocks, and MAFNet~\cite{hong2024mafnet}, which uses attention mechanisms to extract multi-domain features. 

While the majority of prior research on MI classification has centered on EEG, MEG-based models also offer architectural insights relevant to EEG classification. AA-EEGNet~\cite{abdellaoui2020deep}, an attention-enhanced variant of EEGNet, improves cross-subject generalization by emphasizing salient spatiotemporal features. Similarly, DS-GTF~\cite{goene2024dual} fuses Graph Attention Networks with a Transformer encoder to capture spatial and temporal dependencies. Additionally, recently proposed architectures, although developed for other tasks such as sleep staging \cite{kazatzidis2023noveldualstreamtimefrequencycontrastive} and sound localization \cite{kuang2025bast}, offer valuable insights for MI classification, as they share similar architectural and signal processing challenges. Although the aforementioned techniques have demonstrated encouraging classification results, considerable potential for advancement remains. Many recent models still rely on specific preprocessing pipelines or engineered input representations that may limit their adaptability. For instance, 3D-CNN-GAN~\cite{fan2024eeg} requires the transformation of raw EEG signals into temporal-frequency-phase (TFPF) maps, a step that introduces additional complexity and may not generalize well across datasets. Others, like MI-EEGNet~\cite{riyad2021mi}, introduce complex multi-branch architectures that, while effective, increase computational demands and may hinder deployment on resource-constrained devices. Distinctively, our model operates without any preprocessing steps, relying entirely on its architectural design to enhance classification performance. Moreover, the proposed approach prioritizes low latency and computational efficiency, thereby increasing its feasibility for deployment in practical real-time BCI systems.

\section{Proposed Model} \label{sec:propmodel}

\begin{figure*}[t]
    \centering
    \includegraphics[width=\textwidth]{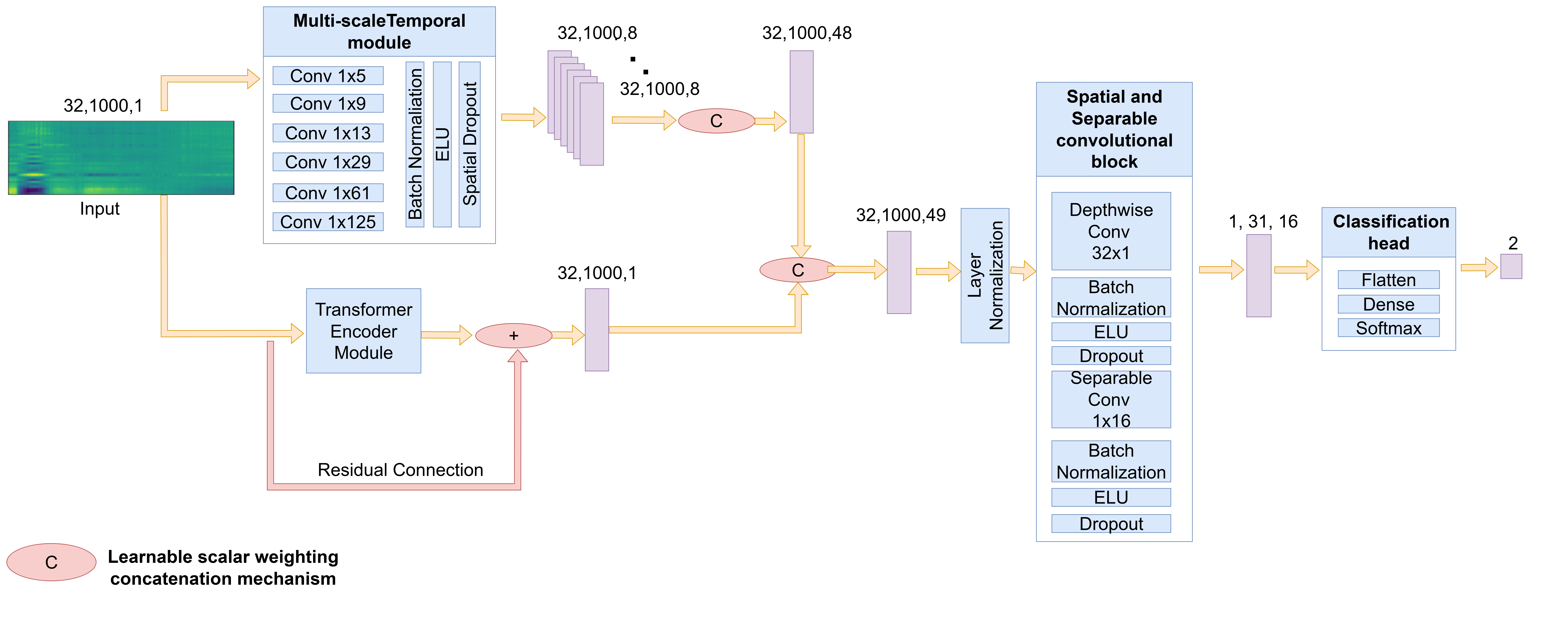}
     \captionsetup{font=footnotesize}
    \caption{\footnotesize EEG-MFTNet's architecture.The numbers displayed above the input and each component’s output (purple boxes) denote the corresponding output dimensions.}
    \label{fig:mymodel}
\end{figure*}

In this work, we propose \textbf{EEG-MFTNet} (EEG-based Multi-Scale Temporal Convolutions and Transformer Fusion Network), a lightweight architecture extending EEGNet \cite{lawhern2018eegnet} with two major enhancements: (1) a multi-scale temporal convolution block inspired by MI-EEGNet \cite{riyad2021mi} and EEG-ITNet \cite{salami2022eeg}, and (2) a Transformer encoder branch, motivated by the dual-stream design in \cite{goene2024dual}. These components operate in parallel during early temporal processing and are fused to provide a richer feature representation, as shown in Figure~\ref{fig:mymodel}.

Each EEG trial is represented as \( X \in \mathbb{R}^{C \times T \times 1} \), where \( C \) is the number of electrodes and \( T \) the number of time samples (e.g., 32 and 1000 in the SHU dataset). The signal is processed through two parallel branches:

\begin{itemize}
    \item \textbf{Multi-Scale Temporal Convolutions:} To capture patterns at varying temporal resolutions, we use six parallel convolutional branches with kernel sizes \( k \in \{5, 9, 13, 29, 61, 125\} \). Each branch uses a \( (1 \times k) \) kernel, outputs 8 feature maps, and incorporates batch normalization, ELU activation, and spatial dropout (rate = 0.5). The outputs of all six branches (48 channels total) are concatenated and scaled with learnable weights, enabling the model to adaptively emphasize the most informative temporal scales.

    \item \textbf{Transformer Encoder:} In parallel, the input is reshaped to \( \mathbb{R}^{T \times C} \), treating each time step as a token and the electrodes as features. A single Transformer encoder block, comprising two attention heads, a feedforward layer (dim = 32), GELU activations, dropout (0.2), and a residual connection, captures long-range dependencies across time. The output is reshaped back to \( \mathbb{R}^{C \times T \times 1} \) for fusion with the convolutional branch.
\end{itemize}

\textbf{Fusion:} The outputs of both branches are merged using trainable scalar weights and concatenated along the channel axis, yielding a fused representation of shape \( 32 \times 1000 \times 49 \). Layer normalization is subsequently applied to enhance training stability.

\textbf{Spatial Processing:} Next, spatial relationships across electrodes are modeled using a depthwise convolution with a kernel size of \( C \times 1 \) and a depth multiplier of 2, yielding 96 output channels. This is followed by batch normalization, ELU activation, average pooling \( (1 \times 4) \), and dropout (rate = 0.3). A subsequent separable convolution \( (1 \times 16) \), consisting of a depthwise and a pointwise component, generates 16 feature maps. This is again followed by batch normalization, ELU activation, average pooling \( (1 \times 8) \), and dropout (rate = 0.3).

\textbf{Classification Head:} Finally, the output is flattened and passed through a dense layer with a max-norm constraint of 0.25, generating class logits. A softmax layer then computes class probabilities, as shown in (1):
\begin{align}
\hat{y} \in \mathbb{R}^N, \quad \sum_{i=1}^{N} \hat{y}_i = 1,
\end{align}
where \( N = 2 \) for binary MI classification. All hyperparameters, such as kernel sizes, number of filters, and dropout rates, were optimized through empirical search, with model selection and checkpointing performed based on validation-set performance.

\section{Data Description}
\label{Data description}
This study uses the SHU dataset~\cite{ma2022large}, a large-scale, publicly available EEG dataset containing data from 25 healthy subjects, each completing five sessions across multiple days. In each session, participants performed 100 trials of left- or right-hand motor imagery, each lasting 4 seconds. EEG signals were recorded from 32 electrodes at 250 Hz, yielding 1000 time samples per trial, which are used directly as input to the model. The dataset was preprocessed by the original authors, who removed baseline drifts and applied a 0.5–40 Hz FIR band-pass filter to retain MI-relevant frequency components. The SHU dataset was selected for its emphasis on session-to-session variability, an ongoing challenge in EEG-based BCIs due to non-stationary signals influenced by factors such as electrode shifts, fatigue, and mental state.

\section{Experimental Results and Discussion}
\label{Experiment Results and Discussion}

\subsection{Experimental Setup}

The proposed model was evaluated on the SHU dataset using the subject-dependent cross-session protocol proposed by the dataset authors~\cite{ma2022large}. Each subject’s first session was used for training (with 20\% held out for validation), while sessions 2–5 were used for testing. A separate model was trained per subject. Experiments were conducted on an NVIDIA GTX 1650 GPU.

To enhance model performance, several training strategies were employed. A model checkpoint callback was used to save the weights corresponding to the highest validation accuracy during training. These best weights were automatically restored upon completion of training. The proposed model, along with all baseline models, was trained for 100 epochs with a batch size of 16, using the categorical cross-entropy loss function and the AdamW optimizer (learning rate of 0.001, weight decay of 1e-4). A learning rate scheduler was also utilized, reducing the learning rate by a factor of 0.5 if the validation loss did not improve for 5 consecutive epochs, down to a minimum of 1e-4. This strategy was adopted to prevent overfitting and to accelerate convergence by dynamically adjusting the learning rate throughout training.
The performance of the proposed model was compared against EEGNet, which served as the strongest baseline in the original SHU dataset paper ~\cite{ma2022large} and formed the architectural foundation for the current work. Additional comparisons were made with two recent EEGNet derivatives: AA-EEGNet and EEG-GeNet.

To ensure a fair comparison, all baselines were trained using the same training settings as our proposed model. Additionally, all baselines used EEGNet-specific hyperparameters consistent with those reported in the SHU paper~\cite{ma2022large}, except for AA-EEGNet, where dropout was set to 0.25. For non-EEGNet-specific parameters, we followed the original configurations provided by each model’s authors, with the exception of EEG-GeNet, where the polynomial degree was set to $K = 3$ after empirical tuning. To ensure reproducibility across all experiments, a fixed random seed of 42 was applied throughout. However, it is important to note that comparisons with other SHU-based studies were excluded due to missing source code, unclear training setups, or protocol inconsistencies with the official SHU benchmark. Thus, we restrict evaluation to models with fully available implementations to ensure fair comparison.

\subsection{Results and Comparison with State-of-the-Art}

Table~\ref{tab:merged_model_comparison} summarizes the comparative classification accuracies obtained by various deep learning architectures evaluated on the SHU dataset. The models considered include EEGNet, the leading baseline reported in the original SHU dataset study \cite{ma2022large}, alongside two recently proposed variants, namely AA-EEGNet and EEG-GENet. These architectures, all based on the EEGNet family, provide relevant baselines for evaluating the performance of the proposed EEG-MFTNet model, which also builds upon and extends EEGNet’s architecture.

As indicated in Table~\ref{tab:merged_model_comparison}, the original EEGNet model attains a mean classification accuracy of 53.7\%. Its derivatives, AA-EEGNet and EEG-GENet, exhibit marginal improvements, achieving 54.8\% and 54.7\% respectively. These modest gains reflect the incremental complexity and refined inductive biases introduced within these extended models. In comparison, EEG-MFTNet shows a noticeable improvement, achieving an average accuracy of \textbf{58.9\%}. This represents an absolute increase of over 5\% compared to the baseline EEGNet, and it also outperforms AA-EEGNet and EEG-GENet by 4.1\% and 4.2\%, respectively.

This notable improvement underscores the efficacy of explicitly capturing both short- and long-range temporal dependencies inherent in EEG signals, an aspect often overlooked in more lightweight architectures. While the standard deviation across subjects increases slightly (10.5\%), this may be due to inter-subject variability, where some subjects exhibit more stable cross-session EEG patterns or patterns that are better captured by our proposed model.

\begin{table}[b] 
    \centering
    \renewcommand{\arraystretch}{1.15}
    \setlength{\tabcolsep}{4.5pt}
     \captionsetup{font=footnotesize}
    \caption{\footnotesize Comparison of classification accuracy (averaged across all sessions and subjects), model complexity, and inference latency (per trial) for various EEGNet-based architectures.}
    \label{tab:merged_model_comparison}
    \resizebox{\linewidth}{!}{
    \begin{tabular}{|l|c|c|c|}
        \hline
        \textbf{Model} & \textbf{Accuracy (\%)} & \textbf{Parameters} & \textbf{Latency (ms)} \\
        \hline
        EEGNet & 53.7\,{\small$\pm$}\,6.3 & \textbf{3,274} & \textbf{47.86} \\
        EEG-GENet & 54.7\,{\small$\pm$}\,8.5 & 7,146 & 48.06 \\
        AA-EEGNet & 54.8\,{\small$\pm$}\,8.9 & 517,498 & 48.83 \\
        \textbf{EEG-MFTNet} & \textbf{58.9}\,{\small$\pm$}\,10.5 & 16,096 & 49.63 \\
        \hline
    \end{tabular}
    }
    \vspace{1mm}
    \caption*{\footnotesize The ``$\pm$'' symbol indicates the standard deviation (in percentage points) across subjects to capture inter-subject variability.}
\end{table}

Regarding computational complexity and inference speed, as shown in Table~\ref{tab:merged_model_comparison}, EEGNet remains the most computationally efficient model, with only 3,274 trainable parameters and the lowest average inference latency of 47.86 ms per trial. EEG-GENet introduces a modest increase in complexity, utilizing 7,146 parameters while maintaining a similar latency of 48.06 ms. In contrast, AA-EEGNet significantly increases model size to 517,498 parameters, over 150 times more than EEGNet, yet results in only a slight latency rise to 48.83 ms.

The proposed EEG-MFTNet balances model expressiveness with computational efficiency. It contains 16,096 parameters, substantially fewer than AA-EEGNet, and demonstrates an average inference latency of 49.63 ms per trial. Although this latency slightly exceeds that of the compared models, it remains well within acceptable bounds for near real-time operation, especially given the considerable improvement in classification accuracy observed in Table~\ref{tab:merged_model_comparison}.

In summary, the integration of multi-scale temporal convolutions with a Transformer-based attention mechanism significantly enhances the EEGNet architecture. This fusion yields EEG-MFTNet, setting a new performance benchmark among lightweight, EEGNet-inspired models for cross-session motor imagery classification. Although EEG-MFTNet introduces moderate computational overhead compared to baseline EEGNet variants, it achieves a favorable trade-off by delivering notably enhanced classification performance, thereby justifying the increased complexity for many practical BCI applications.
\subsection{Analysis of the Results}
The mean classification accuracies for Sessions 2, 3, 4, and 5 were 58.4\%, 57.2\%, 61.0\%, and 58.8\%, respectively, revealing moderate cross-session variability with the highest performance observed during Session 4.

\begin{figure}[h]
    \centering
    \includegraphics[width=1.0\linewidth]{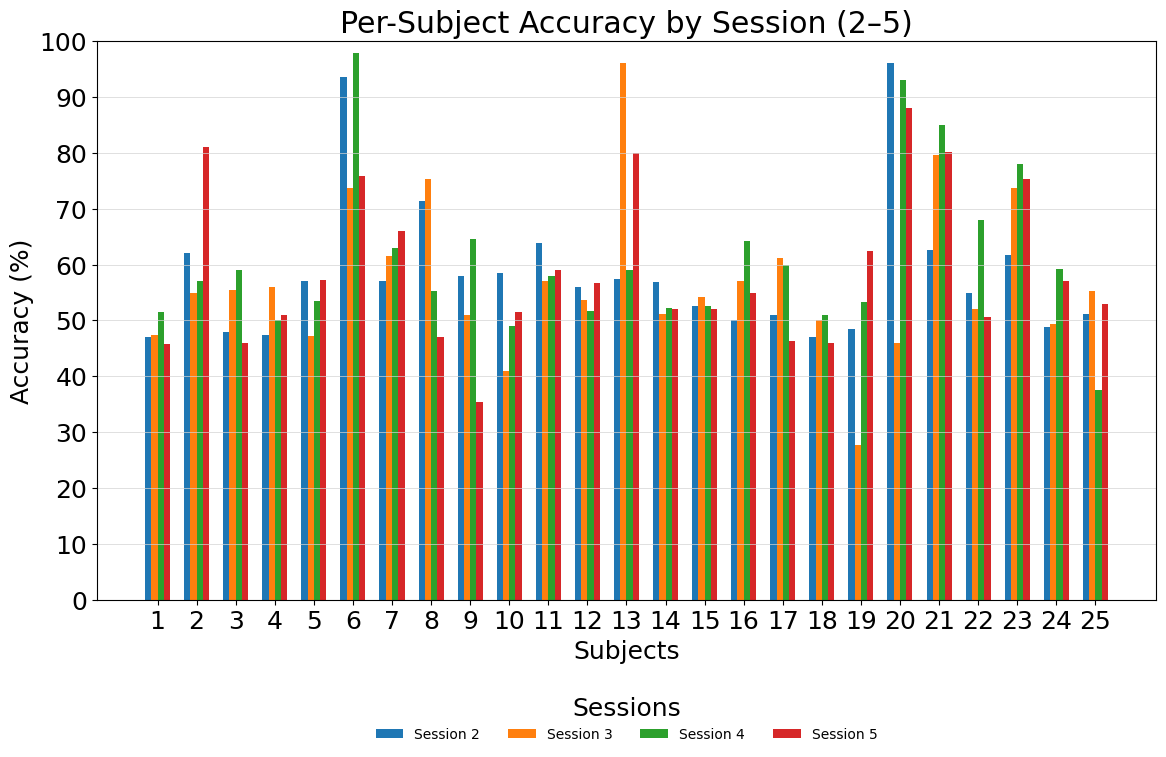}
    \captionsetup{font=footnotesize}
    \caption{ Subject-wise classification accuracy of EEG-MFTNet across the four testing sessions. Each cluster of bars represents a single subject, with each bar indicating the model's accuracy during a specific testing session.}
    \label{fig:Persubject}
\end{figure}

Figure~\ref{fig:Persubject} reveals clear individual differences. Some participants (e.g., Subjects 6 and 20) consistently exceeded 90\% accuracy, suggesting stable motor imagery patterns. Conversely, subjects such as 1, 18, 19, and 25 remained below 55\%, likely due to signal variability or inconsistent imagery. Other subjects (e.g., Subject 13) showed large performance swings across sessions. These results highlight the significant impact of both individual differences and session variability on classification accuracy, underscoring the ongoing challenges in cross-session MI decoding and the need for adaptive approaches.

\subsection{Ablation Study}

To elucidate the individual contributions of the Transformer module and the multi-scale temporal convolution module to the overall efficacy of the proposed model, an ablation study was conducted. This analysis involved systematically removing each component independently and evaluating the resulting impact on classification performance. The findings are summarized in Table \ref{tab:ablation_study}.

As indicated by the results, the exclusion of either the Transformer stream or the multi-scale temporal convolution module caused a marked reduction in classification accuracy. Specifically, the removal of the Transformer component decreased the average accuracy to 57.2\%, whereas omitting the multi-scale temporal module led to a more pronounced decline, with accuracy dropping to 55.2\%. In comparison, the complete model, which integrates both components, attained the highest average accuracy of \textbf{58.9\%}. Despite the performance drops, both ablated variants still outperformed the baseline EEGNet model (53.7\% accuracy). Specifically, adding only the multi-scale temporal module improved accuracy by 3.5\%, while the Transformer encoder alone yielded a 1.5\% gain. These results highlight the complementary contributions of both modules and their combined impact on performance. Even when used individually, each component provides a clear advantage, confirming the effectiveness of the architectural enhancements in EEG-MFTNet.

\begin{table}[b] 
    \centering
    \renewcommand{\arraystretch}{1.2}
    \setlength{\tabcolsep}{8pt}
     \captionsetup{font=footnotesize}
    \caption{ Ablation study of EEG-MFTNet. Checkmarks (\checkmark) indicate the inclusion of the corresponding component. ``--'' in both columns corresponds to the baseline EEGNet.}
    \label{tab:ablation_study}
    \resizebox{\linewidth}{!}{%
    \begin{tabular}{|c|c|c|}
        \hline
        \textbf{Transformer Encoder Stream} & \textbf{Multi-scale Conv Module} & \textbf{Accuracy (\%)} \\
        \hline
        -- & -- & 53.7\textsuperscript{*} \\
        -- & \checkmark & 55.2 \\
        \checkmark & -- & 57.2 \\
        \checkmark & \checkmark & \textbf{58.9} \\
        \hline
    \end{tabular}
    }
    \vspace{1mm}
    \caption*{\footnotesize \textsuperscript{*}EEGNet accuracy reproduced using hyperparameters from the SHU dataset source code~\cite{ma2022large}.}
\end{table}

\section{Model Interpretation}
\label{Model Interpretation}

To gain insight into the model's decision-making process, the Gradient × Input technique was applied to assess the contribution of each EEG channel. This approach has proven effective across diverse deep learning architectures, particularly in EEG-based BCI systems \cite{cui2023towards}. Subject 6, Session 4 was selected for this analysis, as it yielded the highest classification accuracy. Since checkpointing selects the weights with the highest validation accuracy and Subject~6’s training converges rapidly, the resulting predictions tend to remain close to 50\% confidence. Therefore, the model was retrained for 50 epochs on Subject~6’s data using the final weights, yielding more distinct confidence scores and enabling a clearer trustworthiness analysis.

The interpretability method involved computing the element-wise product of the input and the gradient of the model's output with respect to the input. This yielded attribution maps for each correctly classified trial, which were then averaged across time to obtain a single attribution score per channel. Channels were ranked by the average absolute attribution values, highlighting those most influential to the model's decisions.

\begin{figure}[t]
    \centering
    \includegraphics[width=1.\linewidth]{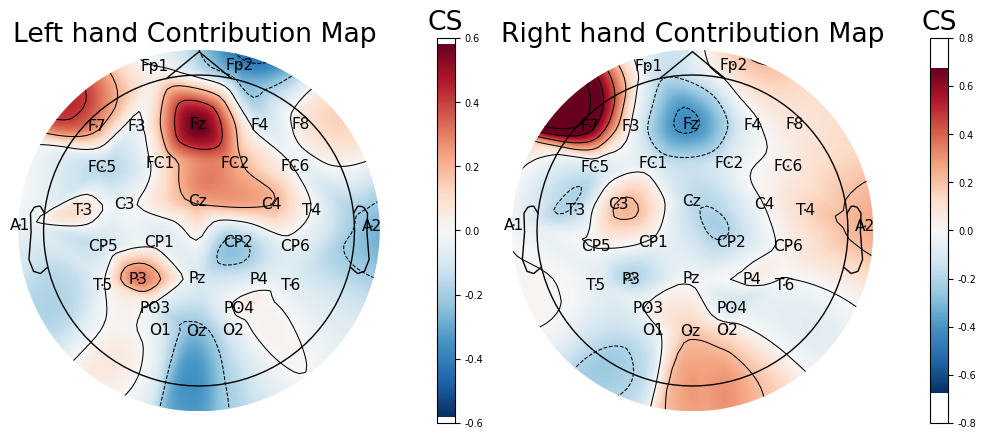}
    \captionsetup{font=footnotesize}
    \caption{ Topographic contribution maps (Gradient × Input) averaged over all correctly classified trials belonging to a specific class (left or right motor representation) for Subject 6, Session 4. Warm colors indicate greater channel importance.}
    \label{fig:sculpaggregatedplots}
\end{figure}

Scalp topographies were generated using MNE-Python to visualize spatial patterns of channel importance, aggregating attribution maps across trials per class. Figure~\ref{fig:sculpaggregatedplots} shows that left- and right-hand imagery produce distinct activation patterns. Notably, right-hand motor imagery is associated with decreased activation in central frontoparietal regions (Fz, F4, FC2, Cz, and FC1) and increased activity in posterior areas (PO3, O1, Oz, and O2), whereas left-hand imagery exhibits the opposite pattern. These observations suggest the model distinguishes classes based on both the presence and suppression of activity in specific regions.

To assess the reliability of the interpretation method, electrode deletion tests were conducted. Electrodes were ranked according to their attribution importance, and increasing proportions of either the most or the least important channels were set to zero. At each step, the model’s prediction confidence was recorded.

\begin{figure}[t]
    \centering
    \subfigure[Left-Hand Class]{
        \includegraphics[scale=0.16]{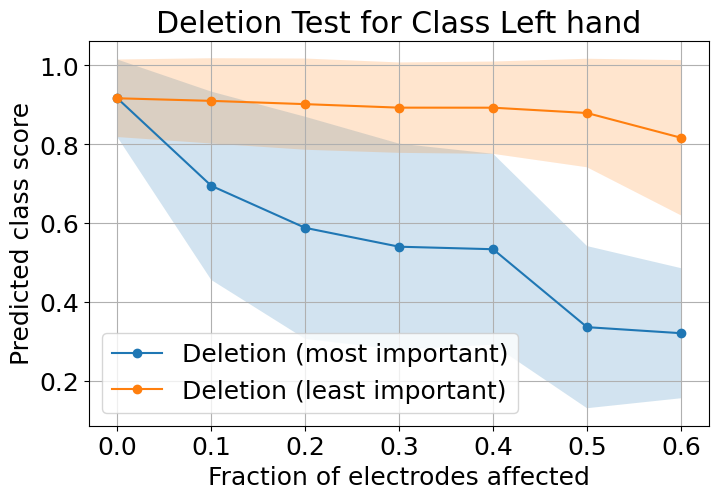}
        \label{fig:left_hand_deletion}
    }
    \hfill
    \subfigure[Right-Hand Class]{
        \includegraphics[scale=0.16]{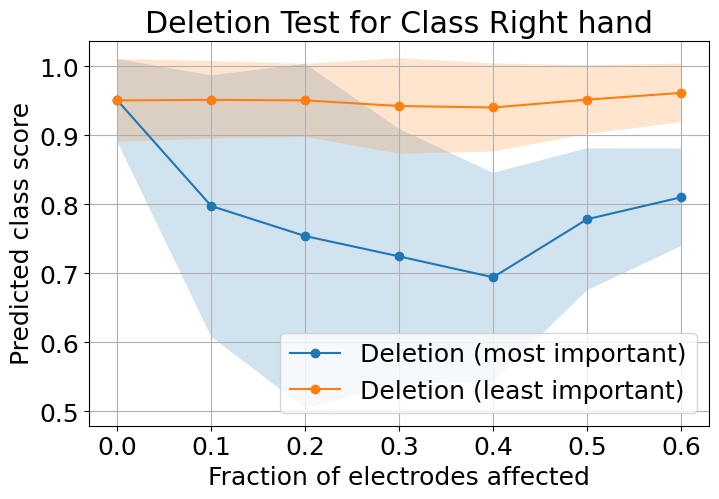}
        \label{fig:right_hand_deletion}
    }
    \captionsetup{font=footnotesize}
    \caption{ EEG-MFTNet's average prediction confidence during Class-specific electrode deletion tests. Blue: most important electrodes deleted; Orange: least important deleted; shaded areas show standard deviation.}
    \label{fig:deletion_tests}
\end{figure}

For left-hand imagery (Figure~\ref{fig:left_hand_deletion}), the confidence dropped from about 0.90 to just below 0.40 after deleting 60\% of the most influential electrodes, while removing the least important channels had only a minor effect. For right-hand imagery (Figure~\ref{fig:right_hand_deletion}), deleting the most important electrodes reduced confidence from nearly 0.95 to around 0.75 at 20\% deletion and to about 0.70 at 40\%, with a slight recovery to approximately 0.80 at 60\%. In contrast, removing the least important channels caused negligible changes.

Overall, the clear separation between the deletion curves obtained by removing the most and the least important electrodes supports the validity of the proposed interpretation method. However, the moderately high area under the curves suggests that, while the method captures the dominant discriminative patterns, it may not fully account for all fine-grained features contributing to the model’s decisions.


\section{Conclusion and Future Work}
\label{Conclusion and Future work}
This work introduced EEG-MFTNet, a deep learning model for motor imagery EEG classification. Extending EEGNet, it integrates a multi-scale temporal convolutional block and a parallel Transformer encoder to capture frequency-specific patterns and long-range dependencies, addressing cross-session variability. Evaluated on the SHU dataset under a subject-dependent cross-session protocol, EEG-MFTNet achieved 58.9\% accuracy, outperforming EEGNet, AA-EEGNet, and EEG-GENet. Despite slightly higher complexity, it maintained low latency and few parameters. Ablation study confirmed the value of both modules, while interpretability analyses revealed attention to class-specific electrodes. Future work will explore cross-session adaptation, broader datasets, and online learning.

\FloatBarrier
\bibliographystyle{IEEEtran}
\bibliography{references}

\end{document}